\documentclass[journal]{IEEEtran}

\ifCLASSINFOpdf
  \usepackage[pdftex]{graphicx}
  \graphicspath{{./figs/}}
  \DeclareGraphicsExtensions{.pdf,.jpeg,.png}
\else
\fi
\usepackage[cmex10]{amsmath}
\usepackage{mathtools}
\usepackage{amssymb}
\usepackage{algorithm}
\usepackage{algpseudocode}
\usepackage[table,xcdraw]{xcolor}

\usepackage{array}

\usepackage{booktabs}
\usepackage{multirow}
\usepackage{tabu}

\usepackage{url}
\PassOptionsToPackage{hyphens}{url}\usepackage{hyperref}

\usepackage{biograph}        
\usepackage{afterpage}

\usepackage{cite}
\usepackage{multirow} 
\usepackage{rotating}

\usepackage{float}
\usepackage{color}
\usepackage{soul}

\usepackage{colortbl}

\def\por1{\partial}

\newcolumntype{S}{>{\centering\arraybackslash} m{.4\linewidth} }

\makeatletter
  \newcommand\tinyv{\@setfontsize\tinyv{5pt}{7}}
\makeatother

\newlength{\hspacephantom}
\begin{document}
\DeclareGraphicsExtensions{.pdf,.jpeg,.png}

\title{MUG: A Parameterless No-Reference JPEG Quality Evaluator Robust to Block Size and Misalignment}




\author{Hossein Ziaei Nafchi, Atena Shahkolaei, Rachid Hedjam, \textnormal{and} Mohamed Cheriet, \IEEEmembership{Senior Member,~IEEE} 
\thanks{H. Ziaei Nafchi, A. Shahkolaei and M. Cheriet are with the Synchromedia Laboratory for Multimedia Communication in Telepresence,
\'Ecole de technologie sup\'erieure, Montreal (QC), Canada H3C 1K3 (email: hossein.zi@synchromedia.ca; atena.shahkolaei.1@ens.etsmtl.ca; mohamed.cheriet@etsmtl.ca)}\\
\thanks{R. Hedjam is with the Department of Geography, McGill University, 805 Sherbrooke Street West, Montreal, QC H3A 2K6, Canada (email: rachid.hedjam@mcgill.ca)}}

\markboth{}%
{}
%


\maketitle

\begin{abstract}
In this letter, a very simple no-reference image quality assessment (NR-IQA) model for JPEG compressed images is proposed. The proposed metric called median of unique gradients (MUG) is based on the very simple facts of unique gradient magnitudes of JPEG compressed images. MUG is a parameterless metric and does not need training. Unlike other NR-IQAs, MUG is independent to block size and cropping. A more stable index called MUG$^+$ is also introduced. The experimental results on six benchmark datasets of natural images and a benchmark dataset of synthetic images show that MUG is comparable to the state-of-the-art indices in literature. In addition, its performance remains unchanged for the case of the cropped images in which block boundaries are not known. The MATLAB source code of the proposed metrics is available at https://dl.dropboxusercontent.com/u/74505502/MUG.m and https://dl.dropboxusercontent.com/u/74505502/MUGplus.m.        
\end{abstract}


%
\IEEEpeerreviewmaketitle


\maketitle

\section{Introduction}
\label{sec:intro}

JPEG lossy compression is one of the most common coding techniques to store images. It uses a block based coding scheme in frequency domain, e.g. discrete cosine transform (DCT), for compression. Since $B \times B$ blocks are coded independent of each other, blocking artifacts are visible in JPEG compressed images specially under low bit rate compression. Several no-reference image quality assessment models (NR-IQAs) have been proposed to objectively assess the quality of the JPEG compressed images \cite{impair1997, freq2000, harmonic2000, Wang2000, DCT2001, Wang2002, orientation2004, Perra2005, Jmath2007, NBM2008, eurasip2009, PCM2010, new2012, NJQA, Learning2014, Tchebichef2014, GridSAR}. NR-IQAs do not need any information of the reference image. NR-IQAs are of high interest because in most present and emerging practical real-world applications, the reference signals are not available \cite{SPM2011}. In the following, we will have an overview on NR-IQAs for JPEG compressed images.

In \cite{impair1997} for each block, horizontal and vertical difference at block boundaries are used to measure horizontal and vertical blockiness, respectively. The authors in \cite{freq2000} proposed a blockiness metric via analysis of harmonics. They used both the amplitude and the phase information of harmonics to compute a quality score. Harmonic analysis was also used to model another blockiness metric in \cite{harmonic2000}. 

Wang et. al. \cite{Wang2000} modeled the blocky image as a non-blocky image interfered with a pure blocky signal. Energy of the blocky signal is then used to calculate a quality score. In DCT domain, a metric was proposed in \cite{DCT2001} that models the blocking artifacts by a 2-D step function. The quality score is calculated following the human vision measurement of block impairments. The metric proposed in \cite{Jmath2007} measures blockiness artifact in both the pixel and the DCT domains. In \cite{NJQA}, zero values DCT coefficients within each block are counted and a relevance map is estimated that distinguishes between naturally uniform blocks and compressed uniform blocks. For this end, an analysis in both DFT and DCT domains is conducted. 

Wang et. al. \cite{Wang2002} proposed an efficient metric that measures blockiness via horizontally and vertically computed features. These features are average differences across block boundaries, average absolute difference between in-block image samples, and zero crossing rate. Using a set of subjective scores, five parameters of this model are estimated via nonlinear regression analysis. In \cite{orientation2004}, the edge orientation changes of blocks were used to measure severity of blockiness artifacts. Perra et. al. \cite{Perra2005} analyzed the horizontal, vertical and intra-block sets of $8\times8$ blocks after applying the Sobel operator to the JPEG compressed images.

The difference of block boundaries plus luminance adaptation and texture masking were used in \cite{NBM2008} to form a noticeable blockiness map (NBM). From which, the quality score is calculated by a Minkowski summation pooling. In \cite{eurasip2009}, 1-D signal profile of gradient image is used to extract block sizes and then priodic peaks in DCT domain are analyzed to calculate a quality score. Chen et. al. \cite{PCM2010} proposed a very similar metric.

In \cite{Learning2014}, three features including the corners, block boundaries (horizontal, vertical and intra-block), and color changes, together with the subjective scores are used to train a support vector regression (SVR) model. Li et. al. \cite{Tchebichef2014} measured the blocking artifacts through weighting a set of blockiness scores calculated by Tchebichef moments of different orders.  
 
Lee and Park \cite{new2012} proposed a blockiness metric that first identifies candidates of having blockiness artifacts. The degree of blockiness of these candidates is then used to compute a quality score. Recently a blockiness metric is proposed that performs in three steps \cite{GridSAR}. Block grids are extracted in the spatial domain and their strength and regularity is measured. Afterwards, a masking function is used that gives different weights to the smooth and textured regions.  

The aforementioned indices have at least one of the following drawbacks. They might not be robust to block size and block misalignment (examples are \cite{Wang2002, orientation2004, Perra2005, NJQA, Tchebichef2014, NBM2008}). They are complex (examples are \cite{DCT2001, NJQA, GridSAR, Tchebichef2014, Learning2014, eurasip2009}), or have many parameters to set (\cite{Wang2002, NJQA, GridSAR, eurasip2009, Tchebichef2014, Learning2014}). Indices like NJQA \cite{NJQA} and GridSAR \cite{GridSAR} are too much slow. Some indices need training (\cite{Wang2002, Learning2014}). Also, the range of quality scores provided by some of the indices like \cite{Wang2002} is not well defined, or they show other numerical issues \cite{GridSAR}.    

In this letter, we propose a quality assessment model for JPEG compressed images that overcomes all aforementioned drawbacks. The proposed index is very simple and efficient, it is parameterless, and robust to block size and misalignment. The proposed metric called MUG is based on two simple facts about blockiness artifact. As a result of more JPEG compression, the number of unique gradient magnitude values decreases, and the median value of unique gradient magnitude values increases. The proposed blockiness metric MUG uses these two simple facts to provide accurate quality predictions for JPEG compressed images. Unlike other metrics that presume position of blocks beforehand or localize the position of blocks, MUG is not a local model and hence does not need any information on the position of blocks.             


\section{Proposed Metric (MUG)}
\label{method}

The proposed index called MUG predicts the quality of JPEG compressed images as follows. Given the JPEG distorted image $\mathcal{D}$, the Scharr gradient operator is used to approximate horizontal $G_x$ and vertical $G_y$ gradients of $\mathcal{D}$: $G_x = h_x \ast \mathcal{D}$ and $G_y = h_y \ast \mathcal{D}$, where $h_x$ and $h_y$ are horizontal and vertical gradient operators, and $\ast$ denotes the convolution. From which, the gradient magnitude is computed as $G(\textbf{x})=\sqrt{G^2_x(\textbf{x})+G^2_y(\textbf{x})}$. It is worth to mention that within the context of the proposed metrics, the Scharr operator performs better than the Sobel and Prewitt operators. The proposed metric works directly on the gradient magnitude instead of directional gradients. Let's denote $uG$ as the unique numerical values of $G(\textbf{x})$. We show in the following that two properties of $uG$ can be used to predict quality of JPEG compressed images: i) number of values in $uG$, and ii) median of $uG$ values.

\subsection{Number of unique gradients (NUG)}
\label{NUG}

The number of unique gradients (NUG), e.g. the number of values in vector $uG$, indicates how many distinct edge strengths exist in JPEG compressed image $\mathcal{D}$. It is very likely that a JPEG compressed image with blocking artifacts has smaller values of NUG than its uncompressed version. To verify this statement, JPEG compressed images of TID2013 dataset were chosen. For each of the 25 distortion-free images in TID2013, there are five JPEG compressed images of different distortion levels. The values of NUG for each of the 25 sets are found inversely proportional to the amount of distortion:

\begin{equation}
  \ \text{Compression rate} \propto \frac{1}{\text{NUG}}
  \label{equ:NUG}
\end{equation}                    

\begin{figure}[htb]
\scriptsize
\begin{minipage}[b]{0.99\linewidth}
  \centering
  \centerline{\includegraphics[height=3.5cm]{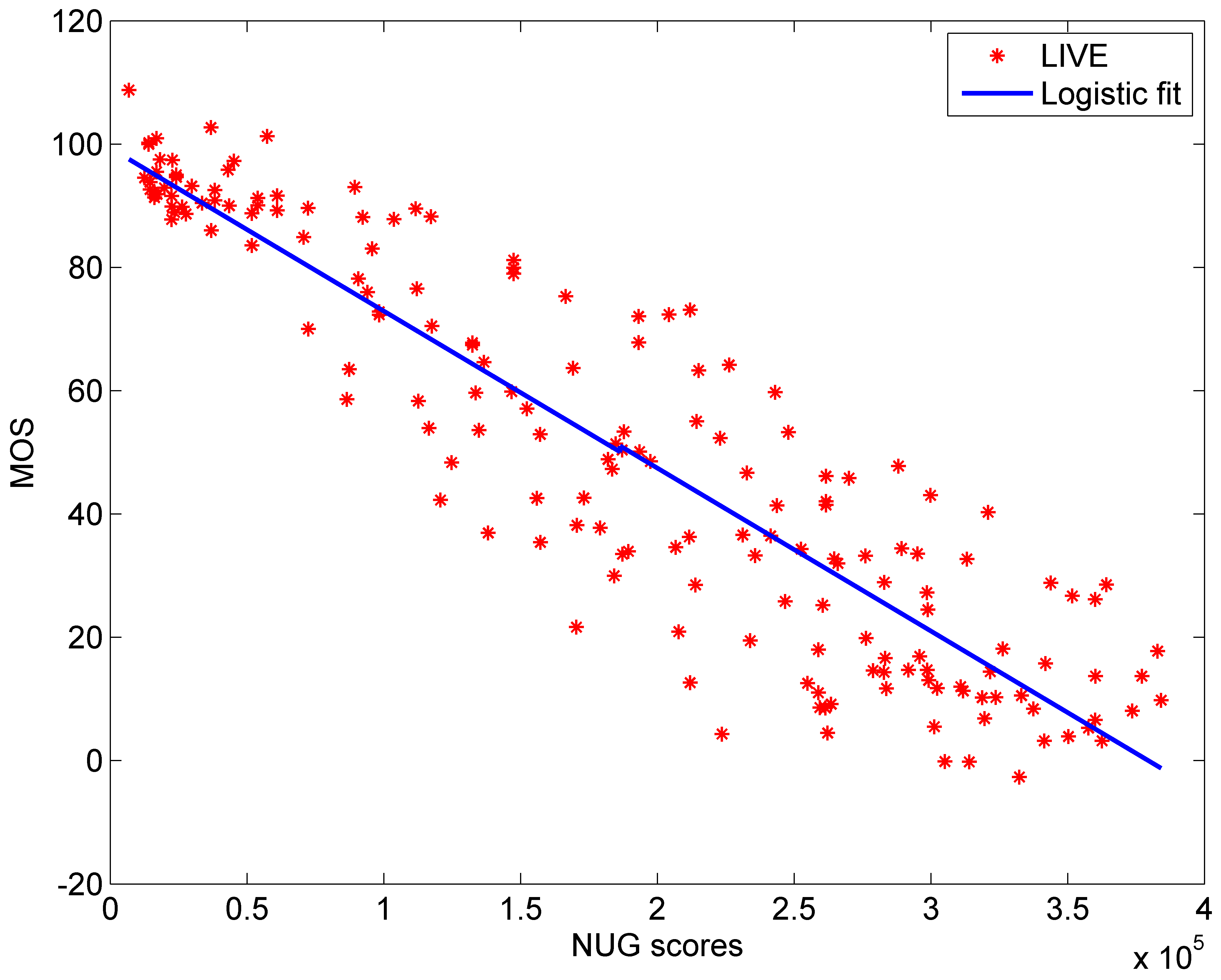}}
\end{minipage}
\caption{Scatter plot of NUG scores against the subjective MOS on the LIVE dataset. The Pearson linear Correlation Coefficient (PLCC) is equal to 0.9105.}
\label{fig1}
\end{figure}

In other words, the Spearman Rank-order Correlation coefficient (SRCC) between NUG values and mean opinion score (MOS) values is equal to 1 for each of the 25 sets. This experiment shows that aforementioned statement holds true. Fig. \ref{fig1} shows scatter plot of NUG scores against the subjective MOS on the LIVE dataset \cite{LIVE} (see experimental results section to see how this plot is drawn). This plot shows that there is noticeable correlation between NUG scores and MOS on this dataset. Unfortunately, NUG does not take into account the content of original images. An image may originally have less edge strengths variation than another. Therefore, there are cases that NUG can not fairly judge images having different contents. This issue is solved through including median of unique gradients ($MUG$) into the proposed model.

\subsection{Median of unique gradients (MUG)}
\label{MUG}

As mentioned above, the image content is a factor that needs to be taken into account. Let's repeat the same experiment on JPEG compressed images of the TID2013 dataset, but this time for the median of unique gradients ($MUG$). The experiments show that the same statement holds true, e.g. the values of $MUG$ for each of the 25 sets are proportional\footnote{Except for one case where SRCC is equal to 0.6, not 1.} to the amount of distortion:

\begin{equation}
  \ \text{Compression rate} \propto {MUG}
  \label{equ:MUG}
\end{equation}                    

In fact, $MUG$ determines how strong is the middle value of unique gradients which helps in taking into account the content of images. However, the values of $MUG$ are not always reliable because image quality is not only related to the edge strengths. The distribution of the unique gradients $uG$ is another factor that can not be considered by direct median value. Therefore, a simple standard deviation normalization was applied on the $uG$ values before median value being computed:

\begin{equation}
  \ uG' = \frac{uG}{\sqrt{ \sigma(uG) }}
  \label{equ:uG}
\end{equation}                    

\begin{figure}[htb]
\scriptsize
\begin{minipage}[b]{0.49\linewidth}
  \centering
  \centerline{\includegraphics[height=3.4cm]{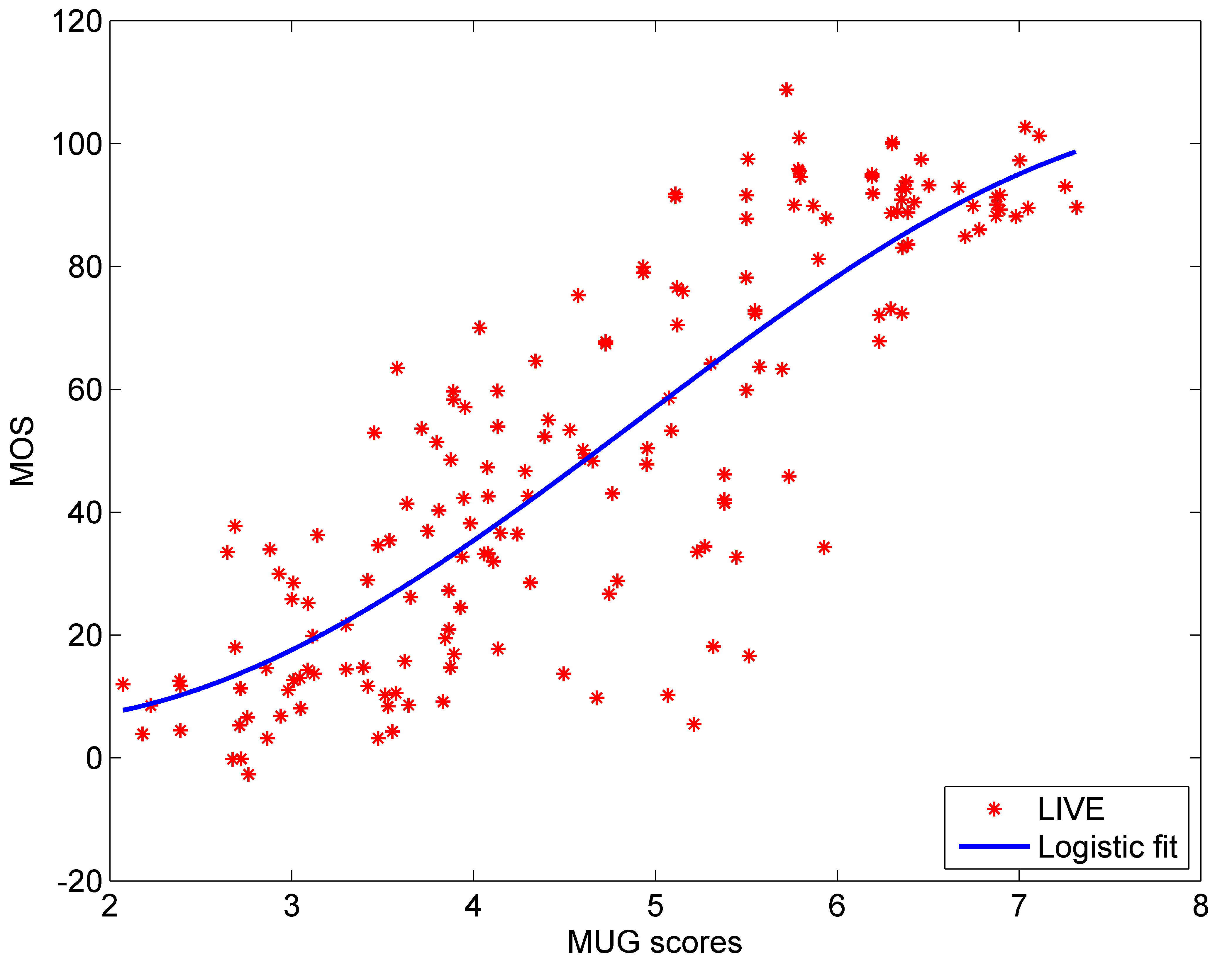}}
\end{minipage}
\begin{minipage}[b]{.49\linewidth}
  \centering
  \centerline{\includegraphics[height=3.4cm]{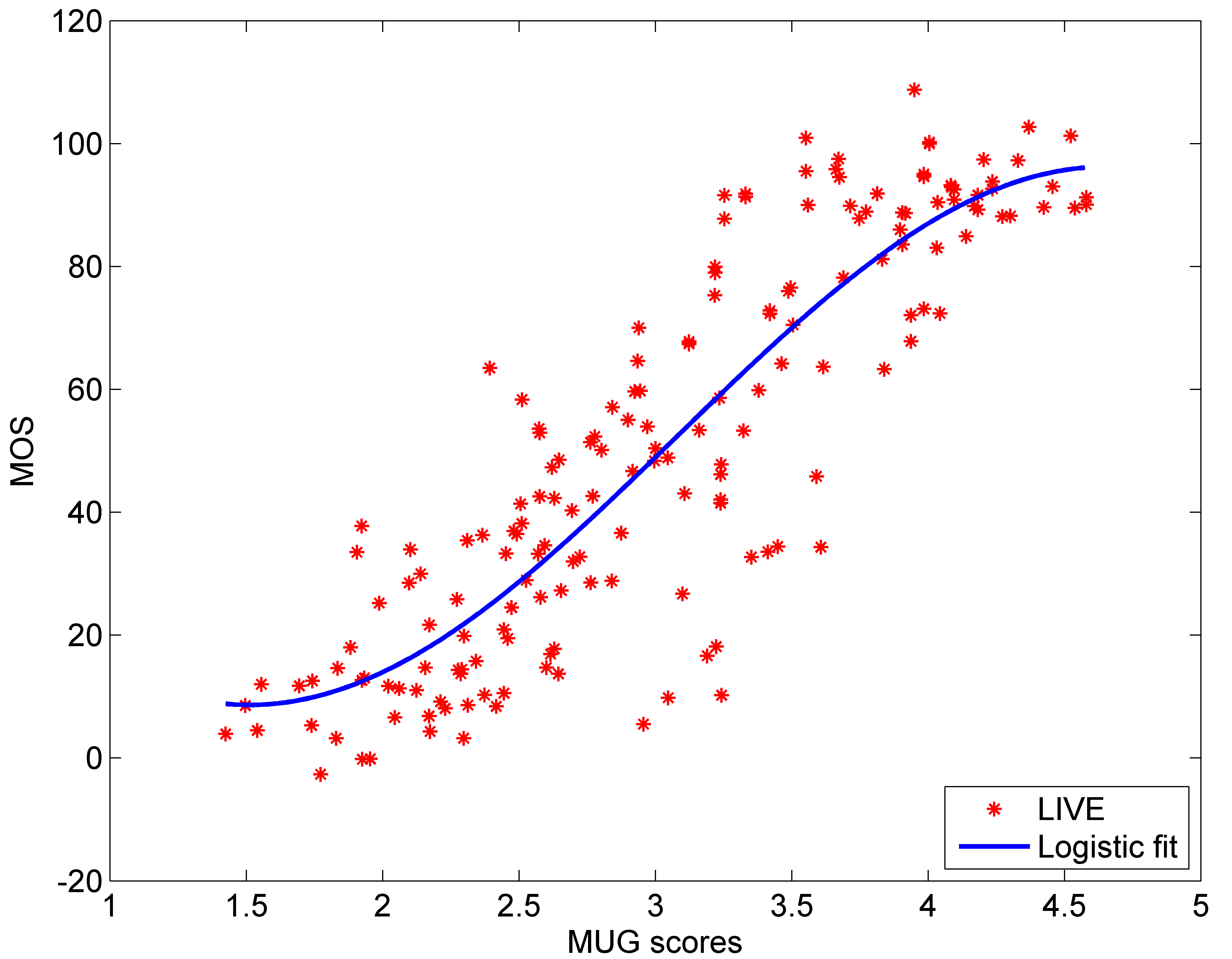}}
\end{minipage}
\caption{Scatter plots of $MUG$ scores against the subjective MOS on the LIVE dataset. Left: $MUG$ without normalization (PLCC = 0.8422), and right: $MUG$ with standard deviation normalization (PLCC = 0.8768).}
\label{fig2}
\end{figure}

Unique gradients vector $uG$ has different behavior for images having naturally uniform regions and block uniform regions. For images with mostly naturally uniform regions, the standard deviation in general decreases by more compression. In contrast, standard deviation value in general increases by more compression for images having less naturally uniform regions. Therefore, median of $uG'$ takes into account the \textit{content} of images. The effect of standard deviation normalization is visually shown in the scatter plots of Fig. \ref{fig2}.

The proposed quality assessment model for JPEG compressed images (MUG) can be written by combining relations (\ref{equ:NUG}) and (\ref{equ:MUG}):

\begin{equation}
  \ \text{MUG} = \frac{MUG}{\text{NUG}}
  \label{equ:Q}
\end{equation}                    

where, $MUG$ (in italic) is the median value of $uG'$. It can be seen that the proposed metric is parameterless. To the best of our knowledge, MUG is the only parameterless metric in the literature. MUG is therefore completely independent to the misalignment. This advantage is shown in the experimental results. Since the proposed metric is parameterless, it should be invariant to the block size as well. However, no dataset is available to experimentally verify this statement. It is worth to mention that when the input image is in color, MUG converts it to a luminance channel: L = 0.06R + 0.63G + 0.27B. According to \cite{invariance2001}, this conversion may be imperfect, but it is likely to offer accurate estimates of differential measurements. Therefore, image gradient computation from L should yield more accurate results. Since MUG only uses the median value of unique gradients, it might not be very accurate for some cases. In the following, MUG is modified by adding a few more unique gradient values.

\subsection{Stable MUG (MUG$^+$)}
\label{stability}

The distributions of unique gradient values can be very different for images having diverse edge information. This distribution might be skewed (usually right-skewed), bimodal, etc. Median value can suddenly be shifted to the left or right if some adjacent values of median are not considered. Therefore, the MUG index can become more stable by adding a few more values to the median. These values must be smaller than the median value because larger values than median have much more variations and might be unreliable. Suppose that $uG'$ values are sorted from smallest to largest. In this case NUG/2 is the index of median value in $uG'$. One easy way to add a few values that mentioned above is to use corresponding values of these indices: NUG$/i$, $i\in \{2, 3, ..., M+1\}$, where $i=2$ is the index of median and $M$ is the total number of values used ($M=19$ in this paper). In fact by adding these extra values, the proposed metric becomes numerically more stable. Moreover, there are cases that there are not $M$ unique values in the vector $uG'$. This property often happens when the majority of the input image or the whole image is naturally uniform or textured. Suppose that there are $1 \leq N \leq M$ of such values available. The stable MUG (called MUG$^+$) takes into account this behavior by the following formulation:

\begin{equation}
  \ \text{MUG}^+ = \frac{\text{MUG}}{M-N+1}
  \label{equ:Q}
\end{equation}                    
where MUG$^+ = $ MUG for $N=M$.

\begin{figure}[htb]
\scriptsize
\begin{minipage}[b]{0.99\linewidth}
  \centering
  \centerline{\includegraphics[height=2.9cm]{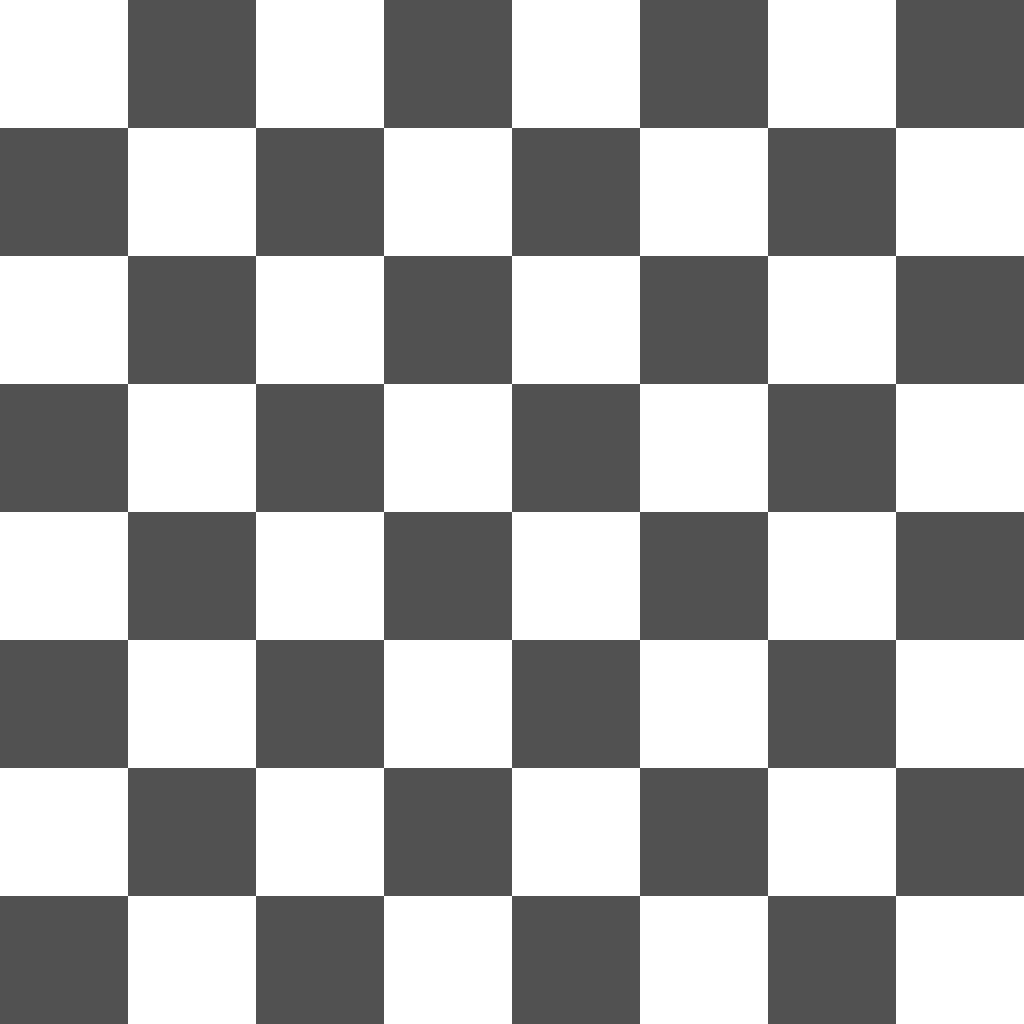}}
\end{minipage}
\caption{A high quality image of chessboard with naturally uniform and textured regions. The image size is 1024$\times$1024 and block sizes are all 128$\times$128.}
\label{chess}
\end{figure}

Apart from the block misalignment problem, several JPEG quality assessment models like \cite{Wang2002, NJQA} provide quite wrong predictions in special cases that image has large amount of naturally uniform regions and/or it is textured. Fig. \ref{chess} shows a high quality image of chessboard. This image has a very bad quality according to the \cite{Wang2002} (Q = -245.89). NJQA \cite{NJQA} likewise assessed this image as being of bad quality (Q = 0.3414). GridSAR \cite{GridSAR} was not able to provide a numerical value. MUG is equal to 0.8060 (very bad quality) which also provides wrong assessment. In contrast, MUG$^+$ = 0.0448 which truly means that chessboard image has a very good quality. This is another advantage of the proposed index MUG$^+$. Note that the datasets used in this paper do not have any image sample with this behavior.                   

\section{Experimental results}
\label{results}

In the experiments, six standard datasets of natural images and a benchmark dataset of synthetic images are used. The TID2013 \cite{TID2013} dataset contains 125 JPEG compressed images in total. The CSIQ dataset \cite{MAD} has 150, LIVE dataset \cite{LIVE} has 175, VCL dataset \cite{VCL} has 138, and the MICT dataset \cite{MICT} has 84 JPEG compressed images. ESPL dataset \cite{ESPL} is a synthetic dataset which contains 100 JPEG compressed images. The TID2008 dataset \cite{TID2008} is another dataset with 100 JPEG compressed images which is in fact a subset of TID2013.  

For objective evaluation, two evaluation metrics were used in the experiments: the Spearman Rank-order Correlation coefficient (SRCC), and the Pearson linear Correlation Coefficient (PLCC). The SRCC and PLCC metrics measure prediction monotonicity and prediction linearity, respectively.    

\begin{figure}[htb]
\scriptsize
\begin{minipage}[b]{0.49\linewidth}
  \centering
  \centerline{\includegraphics[height=3.4cm]{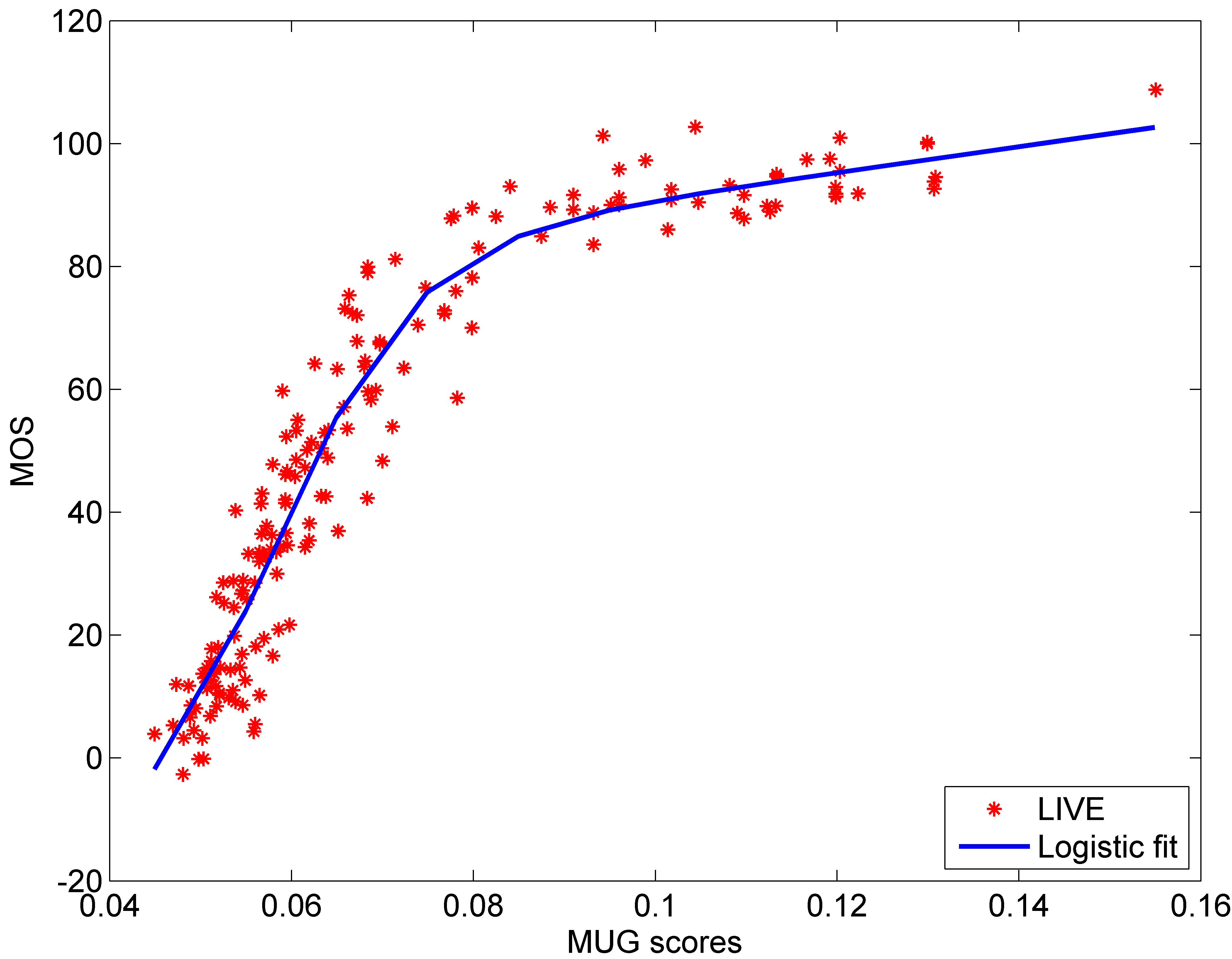}}
\end{minipage}
\begin{minipage}[b]{.49\linewidth}
  \centering
  \centerline{\includegraphics[height=3.4cm]{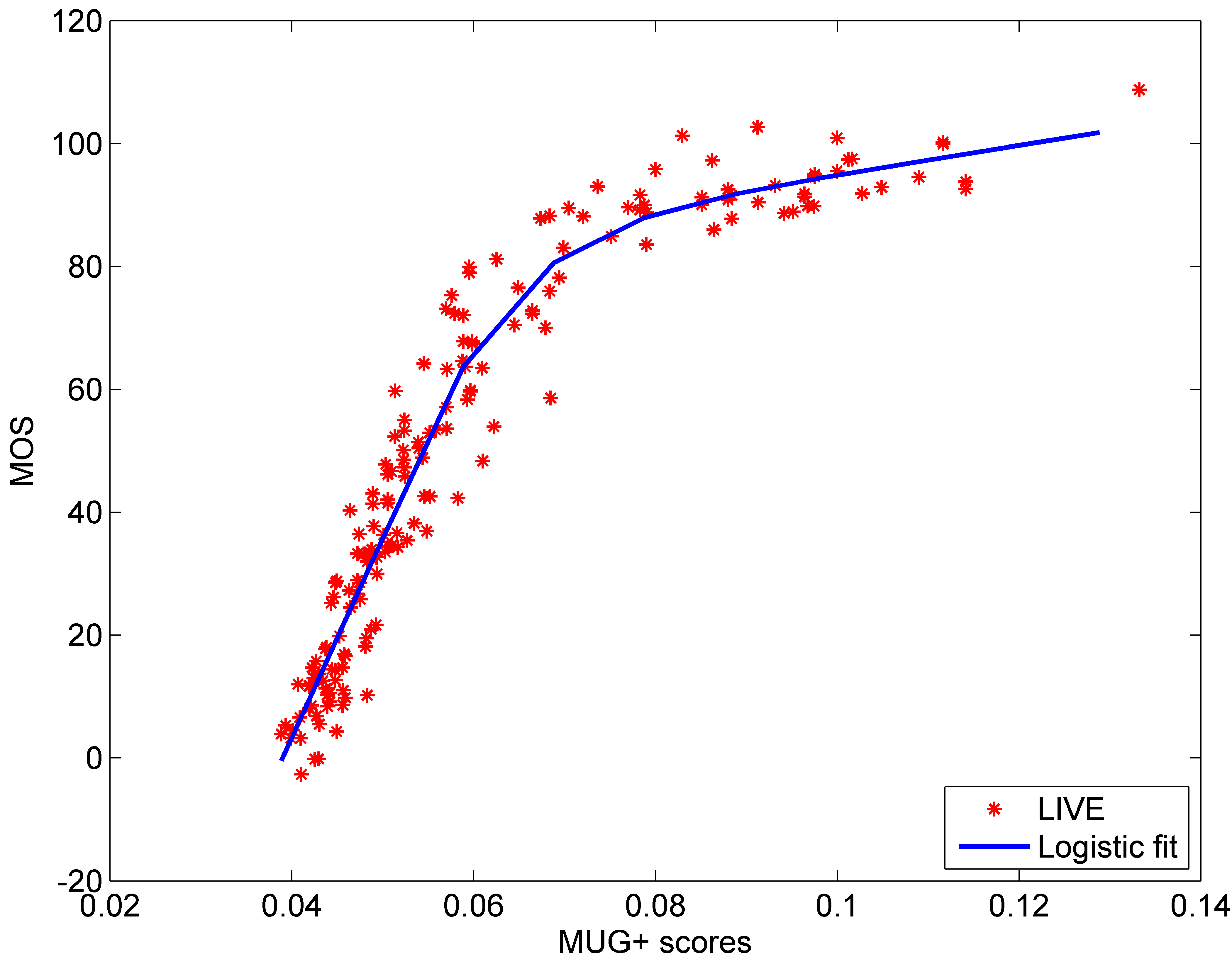}}
\end{minipage}
\caption{Scatter plots of MUG and MUG$^+$ scores against the subjective MOS on the LIVE dataset. Left: MUG (PLCC = 0.9649), and right: MUG$^+$ (PLCC = 0.9730).}
\label{scatter}
\end{figure}

To get a visual observation, the scatter plots of the proposed NR-IQA models MUG and MUG$^+$ on the LIVE dataset are shown in Fig. \ref{scatter}. The logistic function suggested in \cite{LIVE} was used to fit a curve on each plot:

\begin{equation}
  \ f(x) = \beta_1\Big(\frac{1}{2}-\frac{1}{1-e^{\beta_2(x-\beta_3)}}\Big)+\beta_4x+\beta_5
  \label{equ:REG}
\end{equation}                    

where $\beta_1$, $\beta_2$, $\beta_3$, $\beta_4$ and $\beta_5$ are fitting parameters computed by minimizing the mean square error between quality predictions $x$ and subjective scores MOS.

\begin{table}[htb]
\centering
\scriptsize
\caption{PERFORMANCE COMPARISON OF THE IQA MODELS ON JPEG COMPRESSION DISTORTION TYPE OF SEVEN DATASETS IN TERMS OF SRCC AND PLCC}
\begin{tabular}{|l|c|c|ccccc|}
\hline
\multicolumn{2}{|c|}{Index}                                               & SSIM            & \cite{Wang2002}            & NJQA            & \cite{GridSAR}         & MUG    & MUG$^+$            \\
\multicolumn{2}{|c|}{Type}                                                & FR              & NR              & NR              & NR              & NR     & NR              \\ \hline
\multirow{2}{*}{\begin{tabular}[c]{@{}l@{}}TID\\ 2008\end{tabular}} & PLCC & \textbf{0.9540} & 0.9518          & 0.9442          & 0.9511          & 0.9408 & \textbf{0.9529} \\
                                                                    & SRCC & \textbf{0.9252} & 0.9129          & 0.8993          & 0.9166          & 0.9169 & \textbf{0.9239} \\ \hline
\multirow{2}{*}{\begin{tabular}[c]{@{}l@{}}TID\\ 2013\end{tabular}} & PLCC & 0.9544          & 0.9530          & 0.9477          & \textbf{0.9545} & 0.9419 & \textbf{0.9546} \\
                                                                    & SRCC & 0.9200          & \textbf{0.9267} & 0.8860          & \textbf{0.9309} & 0.9077 & 0.9185          \\ \hline
\multirow{2}{*}{CSIQ}                                               & PLCC & \textbf{0.9786} & 0.9751          & 0.9539          & \textbf{0.9788} & 0.9674 & 0.9717          \\
                                                                    & SRCC & 0.9546          & \textbf{0.9551} & 0.9249          & \textbf{0.9565} & 0.9304 & 0.9372          \\ \hline
\multirow{2}{*}{LIVE}                                               & PLCC & \textbf{0.9790} & \textbf{0.9787} & 0.9562          & 0.9756          & 0.9649 & 0.9730          \\
                                                                    & SRCC & \textbf{0.9764} & \textbf{0.9735} & 0.9562          & 0.9726          & 0.9596 & 0.9677          \\ \hline
\multirow{2}{*}{VCL}                                                & PLCC & 0.9257          & \textbf{0.9433} & 0.8611          & \textbf{0.9304} & 0.8683 & 0.8868          \\
                                                                    & SRCC & 0.9236          & \textbf{0.9403} & 0.8445          & \textbf{0.9313} & 0.8659 & 0.8850          \\ \hline
\multirow{2}{*}{MICT}                                               & PLCC & 0.8664          & \textbf{0.8876} & \textbf{0.8746} & 0.8305          & 0.8341 & 0.8503          \\
                                                                    & SRCC & 0.8590          & \textbf{0.8829} & \textbf{0.8728} & 0.8333          & 0.8263 & 0.8513          \\ \hline
\multirow{2}{*}{ESPL}                                               & PLCC & 0.9431          & \textbf{0.9599} & 0.8089          & \textbf{0.9623} & 0.9398 & 0.9370          \\
                                                                    & SRCC & 0.9042          & \textbf{0.9327} & 0.7388          & \textbf{0.9331} & 0.9284 & 0.9265          \\ \hline
\end{tabular}
\label{results1}
\end{table}

SSIM \cite{SSIM} as an FR-IQA, as well as five NR-IQAs including \cite{Wang2002}, NJQA \cite{NJQA}, GridSAR \cite{GridSAR}, and the proposed indices MUG and MUG$^+$ were used in the experiments. \cite{Wang2002} was chosen because it shows outstanding performance, and NJQA because it follows a different approach with promising performance. GridSAR is recently introduced blockiness metric which is also able to handle block misalignment. Table \ref{results1} provides a performance comparison between the six aforementioned FR/NR-IQAs in terms of SRCC and PLCC. The same experiment is repeated on JPEG compressed images with misaligned blocks. JPEG compressed images with misaligned blocks are generated by cutting one pixel from the borders of the images. Since only one pixel width is cropped from image borders, the MOS values should remain unchanged. When block positions are known beforehand, the NR-IQA of \cite{Wang2002} shows the best overall performance for the seven datasets. The proposed indices show consistent prediction accuracy over different datasets and comparable to the GridSAR and SSIM. The proposed indices in general outperform NJQA \cite{NJQA}. When block positions are not known, it can be seen from the Table \ref{results2} that the proposed indices, e.g. MUG and MUG$^+$, and GridSAR show almost the same prediction accuracy as in Table \ref{results1}. This means that they are robust to the block misalignment. In contrast, \cite{Wang2002} provides predictions with low accuracy.

While GridSAR performs better than MUG$^+$ on more considered datasets, it should be noted that GridSAR is a complex metric with several parameters to set. It is also computationally inefficient and numerically unstable.         


\begin{table}[htb]
\centering
\scriptsize
\caption{PERFORMANCE COMPARISON OF THE IQA MODELS ON JPEG COMPRESSION DISTORTION TYPE OF SEVEN DATASETS WITH BLOCK MISALIGNMENT IN TERMS OF SRCC AND PLCC}
\begin{tabular}{|l|c|c|ccccc|}
\hline
\multicolumn{2}{|c|}{Index}                                               & SSIM            & \cite{Wang2002}   & NJQA   & \cite{GridSAR}         & MUG             & MUG$^+$            \\
\multicolumn{2}{|c|}{Type}                                                & FR              & NR     & NR     & NR              & NR              & NR              \\ \hline
\multirow{2}{*}{\begin{tabular}[c]{@{}l@{}}TID\\ 2008\end{tabular}} & PLCC & 0.9247          & 0.3742 & 0.8499 & \textbf{0.9540} & 0.9407          & \textbf{0.9528} \\
                                                                    & SRCC & 0.8989          & 0.3146 & 0.8128 & \textbf{0.9197} & 0.9171          & \textbf{0.9242} \\ \hline
\multirow{2}{*}{\begin{tabular}[c]{@{}l@{}}TID\\ 2013\end{tabular}} & PLCC & 0.9328          & 0.5087 & 0.8540 & \textbf{0.9566} & 0.9418          & \textbf{0.9545} \\
                                                                    & SRCC & 0.9096          & 0.2372 & 0.8107 & \textbf{0.9317} & 0.9075          & \textbf{0.9177} \\ \hline
\multirow{2}{*}{CSIQ}                                               & PLCC & \textbf{0.9750} & 0.6350 & 0.8899 & \textbf{0.9790} & 0.9676          & 0.9718          \\
                                                                    & SRCC & \textbf{0.9504} & 0.5642 & 0.8694 & \textbf{0.9560} & 0.9303          & 0.9370          \\ \hline
\multirow{2}{*}{LIVE}                                               & PLCC & \textbf{0.9761} & 0.5667 & 0.9214 & \textbf{0.9762} & 0.9646          & 0.9728          \\
                                                                    & SRCC & \textbf{0.9722} & 0.4088 & 0.9131 & \textbf{0.9727} & 0.9593          & 0.9673          \\ \hline
\multirow{2}{*}{VCL}                                                & PLCC & \textbf{0.9043} & 0.2949 & 0.6816 & \textbf{0.9265} & 0.8683          & 0.8867          \\
                                                                    & SRCC & \textbf{0.9017} & 0.1923 & 0.6498 & \textbf{0.9268} & 0.8652          & 0.8845          \\ \hline
\multirow{2}{*}{MICT}                                               & PLCC & 0.7967          & 0.4646 & 0.7647 & 0.8189          & \textbf{0.8316} & \textbf{0.8475} \\
                                                                    & SRCC & 0.7865          & 0.4443 & 0.7450 & 0.8217          & \textbf{0.8248} & \textbf{0.8474} \\ \hline
\multirow{2}{*}{ESPL}                                               & PLCC & \textbf{0.9510} & 0.6458 & 0.9414 & \textbf{0.9626} & 0.9398          & 0.9370          \\
                                                                    & SRCC & 0.9144          & 0.6412 & 0.9154 & \textbf{0.9333} & \textbf{0.9285} & 0.9265          \\ \hline
\end{tabular}
\label{results2}
\end{table}

\subsection{Complexity}

To show the efficiency of the proposed indices, a run-time comparison between six IQAs is performed and shown in Table \ref{time}. The experiments were performed on a Corei7 3.40 GHz CPU with 16 GB of RAM. The IQA model was implemented in MATLAB 2013b running on Windows 7. It can be seen that MUG and MUG$^+$ have satisfactory run-times. Compared to the competing metric GridSAR, the proposed metric is about 250 times faster.

\begin{table}[htb]
\centering
\caption{Run time comparison of six IQA models when applied on an image of 1080$\times$1920 size.}
\label{time}
\begin{tabular}{l|c
>{\columncolor[HTML]{656565}}c l|c}
Index & Time (ms) &  & Index    & Time (ms) \\ \hline
JPEGind \cite{Wang2002}   & 140.21      &  & MUG$^+$ & 225.52    \\
SSIM \cite{SSIM}  & 187.85      &  & GridSAR \cite{GridSAR}  & 56810.53    \\
MUG  & 222.06      &  & NJQA \cite{NJQA}     & 79983.76    \\
\hline
\end{tabular}
\end{table}





\section{Conclusion}
\label{conclusion}
In this letter, two novel image quality assessment models for JPEG compressed images were proposed. The proposed indices are very simple and do not need training. They are based on the two simple facts of gradient magnitude of JPEG compressed images. As a result of more
JPEG compression, the number of unique gradient magnitude
values decreases and the median value of unique gradient
magnitude values increases. The extensive experimental results shown that the proposed indices are robust to block misalignment and have consistent performance on seven benchmark datasets.

\section*{Acknowledgments}
The authors thank the NSERC of Canada for their financial support under Grants RGPDD 451272-13 and RGPIN 138344-14.


\bibliographystyle{IEEEtran}
\bibliography{egbib2}   

%






\end{document}